\documentclass[journal]{IEEEtran}

\ifCLASSINFOpdf
\else
   \usepackage[dvips]{graphicx}
\fi
\usepackage{url}

\usepackage{graphicx}
\usepackage{hyperref}
\usepackage{amsmath}
\usepackage{amssymb}
\usepackage{booktabs}
\usepackage{multirow}
\usepackage{tabularx}
\usepackage{arydshln}
\usepackage{makecell}
\usepackage[table]{xcolor}
\usepackage{cite}
\usepackage{float}
\graphicspath{{./}}

\begin{document}

\title{GTransPDM: A Graph-embedded Transformer with Positional Decoupling for Pedestrian Crossing Intention Prediction}

\author{Chen Xie, Ciyun Lin, \IEEEmembership{Member, IEEE}, Xiaoyu Zheng, Bowen Gong, Antonio M. L\'opez, \IEEEmembership{Member, IEEE}
\vspace{-10pt}
\thanks{This work was supported by the Scientific and Technological Developing Project of Jilin Province (20240402076GH). Antonio M. L\'opez acknowledges the financial support to his general research activities given by ICREA under the ICREA Academia Program. Antonio also thanks the Spanish grant PID2020-115734RB-C21 (MCIN/AEI/10.13039/501100011033) for the synergies generated for this paper. Antonio and Chen acknowledge the support of the Generalitat de Catalunya CERCA Program and its ACCIO agency to CVC's general activities. Chen Xie acknowledges the support of the China Scholarship Council (202306170152). \textit{(Corresponding author: Ciyun Lin)}}
\thanks{Chen Xie, Ciyun Lin, and Bowen Gong are with the Department of
Traffic Information and Control Engineering, Jilin University, Changchun
130022, China (e-mail: xiechen19@mails.jlu.edu.cn, linciyun@jlu.edu.cn, gongbowen@jlu.edu.cn).}
\thanks{Xiaoyu Zheng is with the BIT-Barcelona Innovative Transportation Research Group, Civil Engineering School, UPC Barcelona Tech, 08034 Barcelona, Spain (e-mail: xiaoyu.zheng@upc.edu).}
\thanks{Antonio M. L\'opez and Chen Xie are with the Computer Vision Center (CVC), Computer Science Department, Universitat Aut\`onoma de Barcelona (UAB), 08193 Bellaterra, Spain (e-mail: antonio@cvc.uab.es).}}

\maketitle

\begin{abstract}
Understanding and predicting pedestrian crossing behavioral intention is crucial for the driving safety of autonomous vehicles. Nonetheless, challenges emerge when using promising images or environmental context masks to extract various factors for time-series network modeling, causing pre-processing errors or a loss of efficiency. Typically, pedestrian positions captured by onboard cameras are often distorted and do not accurately reflect their actual movements. To address these issues, \textbf{GTransPDM} -- a \textbf{G}raph-embedded \textbf{Trans}former with a \textbf{P}osition \textbf{D}ecoupling \textbf{M}odule -- was developed for pedestrian crossing intention prediction by leveraging multi-modal features. First, a positional decoupling module was proposed to decompose pedestrian lateral motion and encode depth cues in the image view. Then, a graph-embedded Transformer was designed to capture the spatio-temporal dynamics of human pose skeletons, integrating essential factors such as position, skeleton, and ego-vehicle motion. Experimental results indicate that the proposed method achieves 92\% accuracy on the PIE dataset and 87\% accuracy on the JAAD dataset, with a processing speed of 0.05ms. It outperforms the state-of-the-art in comparison.
\end{abstract}

\vspace{-5pt}
\begin{IEEEkeywords}
Pedestrian behavior, Intention prediction, Time-series model, Traffic safety.
\end{IEEEkeywords}

\IEEEpeerreviewmaketitle

\vspace{-10pt}
\section{Introduction}
\begin{figure}[t]
  \centering
  \includegraphics[width=0.5\textwidth]{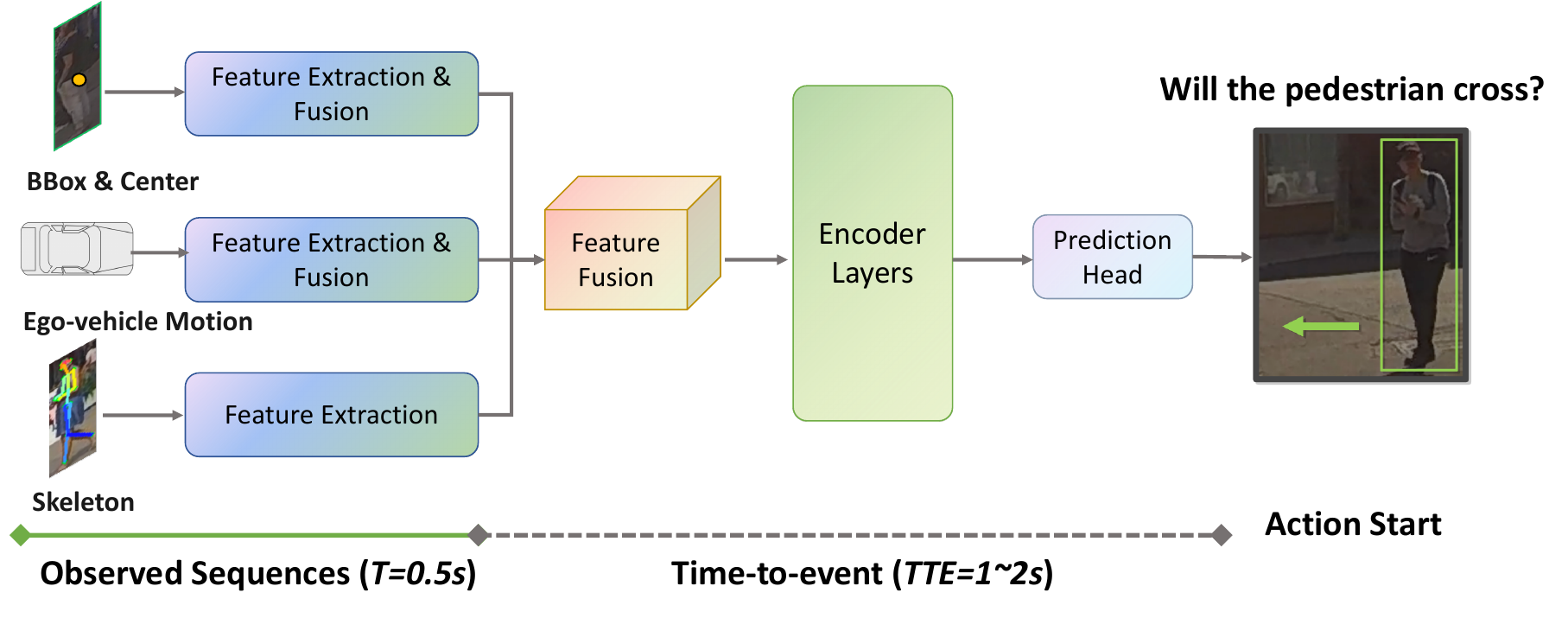}
  \vspace{-20pt}
  \caption{Problem Definition of Pedestrian Crossing Intention Prediction (PCIP): Given $T$ frames of observations, PCIP predicts whether a pedestrian will cross the road within the next 1-2 seconds, allowing the ego-vehicle to react in advance. In this work, we developed a multi-modal fusion approach, incorporating factors such as pedestrian position, human skeleton pose, and ego-vehicle motion.}
  \label{fig1}
\vspace{-20pt}
\end{figure}

\IEEEPARstart{P}{edestrians}, as vulnerable road users, are a crucial concern for the safe operation of autonomous vehicles (AVs). Accurately understanding and predicting pedestrian behavioral intentions is essential for AVs to make proactive safety decisions \cite{8793991,lin2024near}. Pedestrian crossing intention prediction (PCIP) continues to be a challenge in time series classification, where the input comprises a sequence of features derived from sensor data, and the output indicates the probability that a person will cross the road after a given prediction horizon \cite{kotseruba2021benchmark} (Fig. \ref{fig1}).  

In existing research, RNN-based approaches have been employed to develop time series models, either independently or in combination with graph neural networks (GNNs), to represent skeletal information \cite{9743954,10418196,zhang2023trep,ahmed2023multi,8876650}. However, the role of skeleton information within the feature fusion paradigm remains controversial in current studies, underscoring the necessity for precise extraction of pose dynamics \cite{10535049,lorenzo2021intformer}. To gain a deeper insight into pedestrian crossing behavior, various factors have been explored in existing approaches, showcasing improved performance by incorporating environmental context \cite{10161318,9345505,rasouli2020pedestrian,zhao2021action}. However, these methods may lead to errors in image processing or instance mask inference, and may also become computationally intensive. Pedestrian trajectories provide valuable insights into pedestrian behavior and have been extensively utilized in PCIP \cite{10493148,9197434,9827084,9304591}. Yet, since pedestrian positions in the image are relative to the ego-vehicle, this can hinder a model's ability to accurately identify pedestrian movements in real-world scenarios. Typically, a pedestrian's lateral shift toward the ego-vehicle is a key indicator of a potential crossing. However, the position may be distorted due to the motion of the ego-vehicle, making it challenging to pinpoint their actual movements. Moreover, the distance between pedestrians and the ego-vehicle greatly influences the crossing decision \cite{dang2024coupling,soares2021cross}. This factor, though critical, has often been neglected in current studies due to the complexity of depth estimation.

To address the above issues, a \textbf{G}raph-embedded \textbf{Trans}former with a \textbf{P}osition \textbf{D}ecoupling \textbf{M}odule (\textbf{GTransPDM}) was developed by merging multi-modal features. Based on the observation that \textit{crossing pedestrians in conflict with the ego-vehicle inevitably exhibit lateral movement toward it}, we propose a simple yet intuitive PDM to enhance pedestrian lateral movement recognition using reference lines. Then, the area ratio of the bounding boxes was proposed as a proxy representation of depth variation, demonstrating strong consistency with actual depth measurements. Finally, residual graph convolutional blocks with learnable edge importance were designed within a Transformer-based architecture for PCIP, enabling the consideration of spatio-temporal variations in human pose skeletons. The main contributions are summarized as follows.
\begin{itemize}
\item We are the first to address the motion disparity between ego-vehicles and pedestrians to improve pedestrian position representations from on-board sensors. Building on this insight, we introduce PDM, a method that decouples the movements of AVs and pedestrians to enhance the recognition of pedestrians approaching the vehicle. Moreover, we leverage the area ratio of bounding box as depth cues instead of costly depth estimation.
\item We propose a fast Transformer-based multi-modal fusion framework to enhance PCIP by integrating data from various sensors. Notably, we develop residual graph convolutional blocks with learnable edge importance to capture the spatio-temporal dynamics of the human pose skeleton, which allows for accurate modeling of joint interactions and motion patterns.
\item Extensive experiments on the large-scale PIE and JAAD datasets demonstrate the superior performance of our proposed method, achieving an inference time of 0.05 ms using a single GPU, outperforming existing methods.
\end{itemize}

\vspace{-15pt}
\section{Methodology}
\vspace{-5pt}
\begin{figure}[t]
  \centering
  \includegraphics[width=0.5\textwidth]{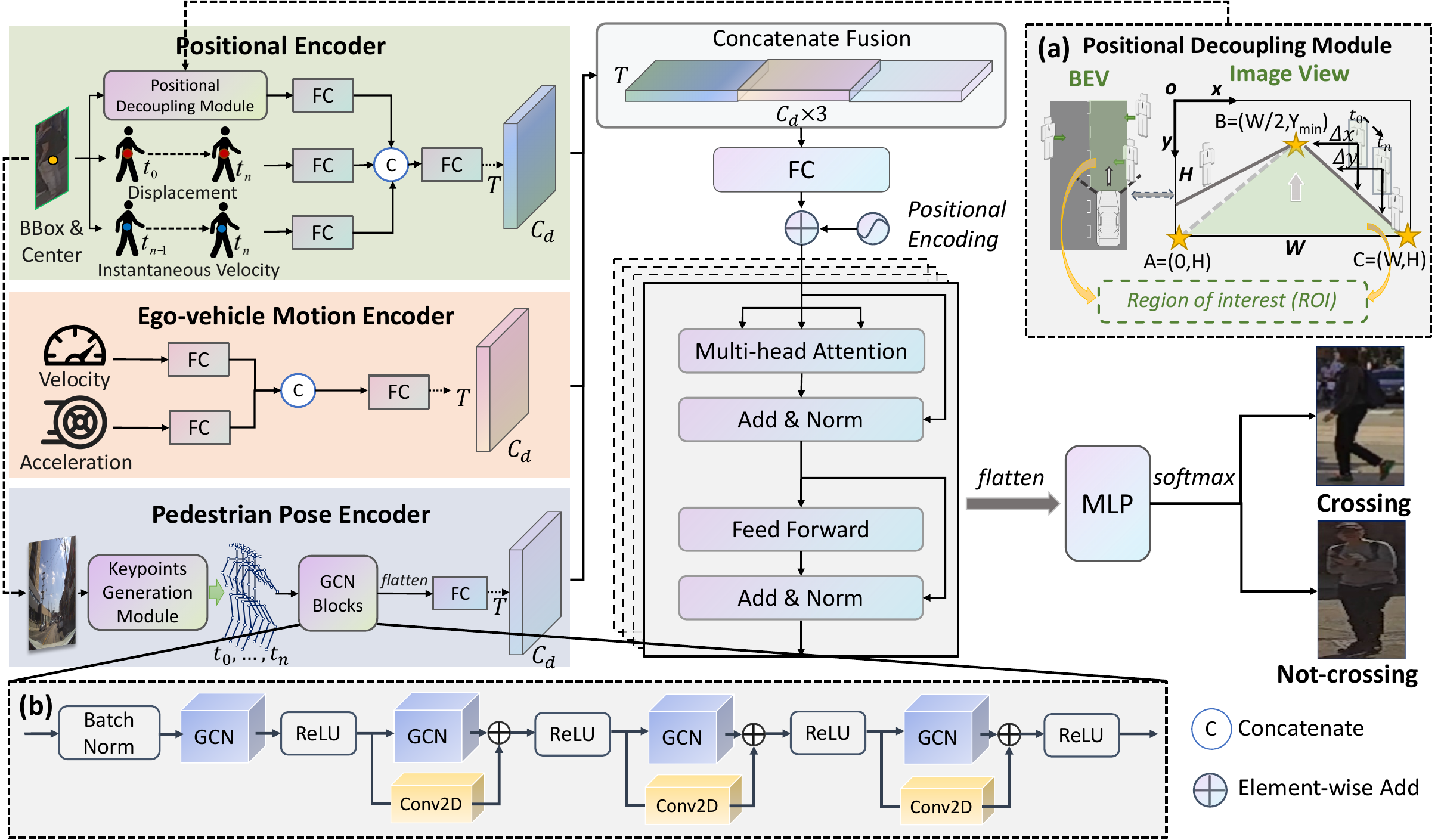}
  \vspace{-20pt}
  \caption{Overview of the proposed framework.}
  \label{fig2}
\vspace{-20pt}
\end{figure}
   
\subsection{Overview}
\vspace{-5pt}

An overview of the proposed framework is presented in Fig.~\ref{fig2}. In the presented framework, the encoding of the network was performed using position, pose, and ego-vehicle motion toward pedestrian $i$ across $T$ observation frames.  Each encoder was referred to as $\mathbf{X}_{pe}$, $\mathbf{X}_{ke}$, and $\mathbf{X}_{ev}$, respectively. In positional encoder, PDM $P_i$ was developed to decompose true lateral movements of the pedestrian and reflect depth changes, which were then integrated with displacement $D_i$ and instantaneous velocity $V_i$ to further represent the positional shift in image view. Then, the ego-vehicle's motion, which indicates potential conflict with a pedestrian, was characterized by its speed $S_i$ and acceleration $Acc_i$. In addition, several Graph Convolutional Network (GCN) blocks equipped with residual connections were layered to capture spatial dynamics of the pedestrian pose $K_i$. Finally, these encoded representations were combined and fed into a Transformer encoder for temporal modeling, followed by an MLP head for classification. As described above, the PCIP problem in this study can be outlined as follows.

\vspace{-15pt}
\begin{equation}
\label{eq1}
\hat{Y}_i=\operatorname{argmax} \mathcal{P}\left(\mathbf{X}_{pe}\left[P_i, D_i, V_i\right], \mathbf{X}_{ke}\left[K_i\right], \mathbf{X}_{ev}\left[S_i, Acc_i\right]\right)
\end{equation}

\vspace{-20pt}
\subsection{Network Architecture}

\textbf{Positional Encoder.} In this study, the positional encoder includes displacement $D_i=\left\{x_t-x_0, y_t-y_0\right\} \in \mathbb{R}^{T \times 2}$, instantaneous velocity $V_i=\left\{x_t-x_{t-1}, y_t-y_{t-1}\right\} \in \mathbb{R}^{T \times 2}$, and $P_i$ from the proposed PDM, where $x_t$ and $y_t$ are the center coordinates of pedestrian bounding boxes, and  $t\in(0,...,T)$. $D_i$ and $V_i$ collectively describe the overall and frame-by-frame displacement of the pedestrian relative to the ego-vehicle. Our proposed PDM is shown in Fig.~\ref{fig2} (a). Firstly, the region of interest (ROI) for the crossing behavior was enclosed using two reference lines $(x_l, y_l)$. Then, the disparity between the pedestrian center and the reference line at frame $t$ was calculated, namely $(\Delta x_t,\Delta y_t)$, where $\Delta x_t=x_t-x_l$, $\Delta y_t=y_t-y_l$ to indicate position disparity between them. Finally, the area ratio $R$ was employed to combine with the position variations in PDM to capture the depth shift in image view. The whole features in $P_i$ can be represented as:

\vspace{-15pt}
\begin{equation}
\label{eq2}
\begin{gathered}
P_i=\left\{\Delta x_t-\Delta x_{t-1}, \Delta y_t-\Delta y_{t-1}, R\right\} \in \mathbb{R}^{T \times 3} \\
R=\left(\frac{A_t}{A_{t-1}}-1\right) \times \alpha
\end{gathered}
\end{equation}
\vspace{-10pt}

where: $A_t$ is the bounding box area at frame $t$. $\alpha$ is a scaling factor and was set as $\alpha=100$. Each feature in $\mathbf{X}_{pe} = [P_i, D_i, V_i]$ was passed through a fully connected (FC) layer, then concatenated along the channel dimension for fusion. Finally, another FC layer was applied for linear projection to generate the final representation:

\vspace{-15pt}
\begin{equation}
\label{eq3}
\mathbf{X}_{pe}=\mathbf{F C}\left(\operatorname{Concat}\left[\mathbf{F C}\left(P_i\right), \mathbf{F C}\left(D_i\right), \mathbf{F C}\left(V_i\right)\right]\right) \in \mathbb{R}^{T \times C_d}
\end{equation}
\vspace{-15pt}

where: $C_d$ indicates the channel dimension of $X_{pe}$.

\textbf{Ego-vehicle Motion Encoder.} The motion of the ego-vehicle indicates a potential conflict with a pedestrian. Thus, speed $S_i=\left\{s_t\right\} \in \mathbb{R}^{T \times 1}$ and acceleration $Acc_i=\left\{acc_t\right\} \in \mathbb{R}^{T \times 1}$ of the ego-vehicle were utilized simultaneously to describe it. To represent the global $Acc_i$, it was calculated as:

\vspace{-15pt}
\begin{equation}
\label{eq4}
a c c_t=\frac{\left(s_T-s_0\right) \times F P S}{3.6 \times T}
\end{equation}
\vspace{-10pt}

where: $FPS$ presents the frames per second depending on video attributes. Furthermore, similar to $\mathbf{X}_{pe}$, the $\mathbf{X}_{ev}$ was also encoded and fused by FC layers and concatenate operation:

\vspace{-15pt}
\begin{equation}
\label{eq5}
\mathbf{X}_{ev}=\mathbf{F C}\left(\text { Concat }\left[\mathbf{F C}\left(S_i\right), \mathbf{F C}\left(A c c_i\right)\right]\right) \in \mathbb{R}^{T \times C_d}
\end{equation}
\vspace{-15pt}

\textbf{Skeleton Pose Encoder.} The pedestrian skeleton pose reflects the action highly related to the crossing behavior, which can be represented as a graph $G=\{V,E\}$ naturally, where the joints are represented as nodes $V$ and limbs as edges $E$. $N=20$ keypoints $K_i=\left\{K_i^1, K_i^2, \ldots, K_i^N\right\}\in \mathbb{R}^{T \times N \times 3}$ was utilized, with each point is represented as $K_i^j=\left\{\left(x_t^{\prime}, y_t^{\prime}, s_t\right) \mid j \in 1, \ldots, N\right\}$. The adjacency matrix $\mathbf{A}$ includes the skeleton naturally connections and self-connected matrix. To avoid redundancy with position, each absolute coordinates of keypoint $(x_t^k,y_t^k)$ in the image was normalized with its corresponding top-left bounding box coordinates $(x_t^{b t l},y_t^{b t l})$ as: $\left(x_t^{\prime}, y_t^{\prime}\right)=\left(x_t^k-x_t^{b t l}, y_t^k-y_t^{b t l}\right)$, and $s_t$ is the confidence score. Inspired by the convolution in graph structure~\cite{kipf2016semi}, four stacked GCN blocks with residual connections were proposed as feature extractors of $K_i$, as shown in Fig.~\ref{fig2} (b). Furthermore, learnable edge importance was also applied in the network training. The entire process can be denoted as: 

\vspace{-15pt}
\begin{equation}
\label{eq6}
\begin{gathered}
\widehat{\mathbf{A}}=\mathbf{D}^{-\frac{1}{2}} \mathbf{A D}{ }^{-\frac{1}{2}} \\
\mathbf{X}_i=\text { BatchNorm }\left(K_i\right) \\
\mathbf{H}_0=\sigma\left((\mathbf{E}_0 \odot \widehat{\mathbf{A}}) \mathbf{X}_i \mathbf{W}_0\right)   \\
\mathbf{H}_l=\sigma\left((\mathbf{E}_l \odot \widehat{\mathbf{A}}) \mathbf{H}_{l-1} \mathbf{W}_l+\operatorname{Conv2D}\left(\mathbf{H}_{l-1} ; \mathbf{W}_{c l}\right)\right) \\
\end{gathered}
\end{equation}
\vspace{-10pt}

where: $\mathbf{D}$ is the diagonal node degree matrix. $\mathbf{E}_l$ represents the learnable edge matrix that align with the dimension of $\mathbf{A}$ in layer $l$ ($l=1,2,3$), while $\mathbf{H}_l$ means the feature map. $\sigma(\cdot)$ indicates the ReLU activate function, and $\odot$ is Hadamard product operation. Further, the last layer $\mathbf{H}_3$ was flattened to $\mathbb{R}^{T \times (N \times d_{hid})}$ and linear projected as the output of $\mathbf{X}_{ke}$:

\vspace{-10pt}
\begin{equation}
\label{eq7}
\mathbf{X}_{ke}=\mathbf{F C}\left(\text { Flatten }\left(\mathbf{H}_3\right)\right) \in \mathbb{R}^{T \times C_d}
\end{equation}
\vspace{-20pt}

where: $d_{hid}$ is the dimension of hidden layers in GCN. After encoder layers, the $\mathbf{X}_{pe}$,$\mathbf{X}_{ev}$,and $\mathbf{X}_{ke}$ were concatenated along the channel dimension for feature fusion, with another FC layer mapping to the Transformer embedding space.

\vspace{-10pt}
\begin{equation}
\label{eq8}
\mathbf{X}=\mathbf{F C}\left(\operatorname{Concat}\left[\mathbf{X}_{pe}, \mathbf{X}_{ev}, \mathbf{X}_{ke}\right]\right) \in \mathbb{R}^{T \times C_d}
\end{equation}
\vspace{-20pt}

So far, $\mathbf{X}$ comprises features from all modalities. To capture long-range dependencies, Transformer encoder layers~\cite{vaswani2017attention} were employed by leveraging the self-attention mechanism to model the temporal dimension. Additionally, positional encoding (PE) was applied to preserve temporal order. After stacking several Transformer encoder layers, the output $\mathbf{X}_{\text{out}}$ was flattened and passed through an MLP head with a single fully connected layer for binary classification.

\vspace{-12pt}
\begin{equation}
\label{eq12}
\hat{Y}_i=\text {softmax}\left(\mathbf{F C}\left(\text { Flatten }\left(\mathbf{X}_{\text {out }}\right)\right)\right)
\end{equation}
\vspace{-10pt}

\vspace{-20pt}
\section{Experiments}
\subsection{Datasets and Implementation Details}
The effectiveness of the proposed method was evaluated using two large-scale datasets in the context of autonomous driving -- JAAD~\cite{8265243} and PIE~\cite{rasouli2019pie}. Both datasets are annotated with the crossing point in time, enabling the TTE calculation. The video frame rate from the two datasets is 30 FPS. Instead of ego-vehicle speed, five ego-vehicle motion states were used in the $\mathbf{X}_{ev}$ of JAAD, as velocity data are not available. The data used in the PIE experiments was split into train, validation (val), and test at a ratio of 0.5:0.1:0.4, with random splitting based on person IDs. For comparison with existing methods (Table \ref{tab1} and \ref{tab2}), the performance following the data split in~\cite{kotseruba2021benchmark} was also evaluated, with Set 01, 02, 04 for training, Set 05, 06 for validation and Set 03 for testing. Each video set contains a continuous driving recording segmented into multiple 10-minute clips. For the JAAD dataset, the JAAD\_all subset and adhered to the default data split were used~\cite{kotseruba2021benchmark}. Furthermore, 0.5 second ($T=16$) observations were utilized to predict crossing behavior after 1-2 seconds ($TTE \in [30,60]$), sampled data based on the sliding window with an overlap of 0.6, 0.8 for PIE and JAAD dataset, respectively.

In the network architecture, the dimension of three encoder layers $C_d$, hidden layers in GCN $d_{hid}$, and hidden size in Transformers are all set to 64. With the concatenate operation in the feature fusion block, the channel dimension of the fused feature is $3 \times C_d=192 $. In the transformer encoder, $dropout=0.1$, 4 attention heads, and 4 stacked layers were applied to extract temporal information. The reference lines in PDM were set to one line between points: $A=(0,H)$ and $B=(W/2,Y_{min})$ while another between: $B=(W/2,Y_{min})$ and $C=(W,H)$, as shown in Fig.~\ref{fig2} (a), where $Y_{min}$ is the minimum $y_t$ of all samples in each dataset. The minor differences in coordinates have little impact on model performance in subsequent experiments, as shown in Table \ref{tab3}. The keypoints were generated by the pre-trained Alphapose~\cite{fang2022alphapose}, while the additional neck, hip, and body center points were further calculated based on the average of adjacent points. 

The cross-entropy loss~\cite{cox1958regression} was utilized for network training. For the PIE dataset, we trained for 32 epochs with a batch size of 128, using the Adam optimizer with an initial learning rate of \(1e^{-4}\), and employed the ReduceLROnPlateau scheduler to halve the learning rate if the validation loss do not decrease by \(1e^{-4}\) over 8 epochs. In JAAD, the AdamW optimizer with a constant learning rate of $5e^{-5}$ was applied for 32 epochs training, with a batch size of 64. Additionally, a weight decay of \(1e^{-4}\) was applied on both datasets to mitigate overfitting. All experiments were conducted on a server equipped with an NVIDIA TITAN X GPU with 12 GB of memory.

\begin{figure*}[t]
  \centering
  \includegraphics[width=1\textwidth]{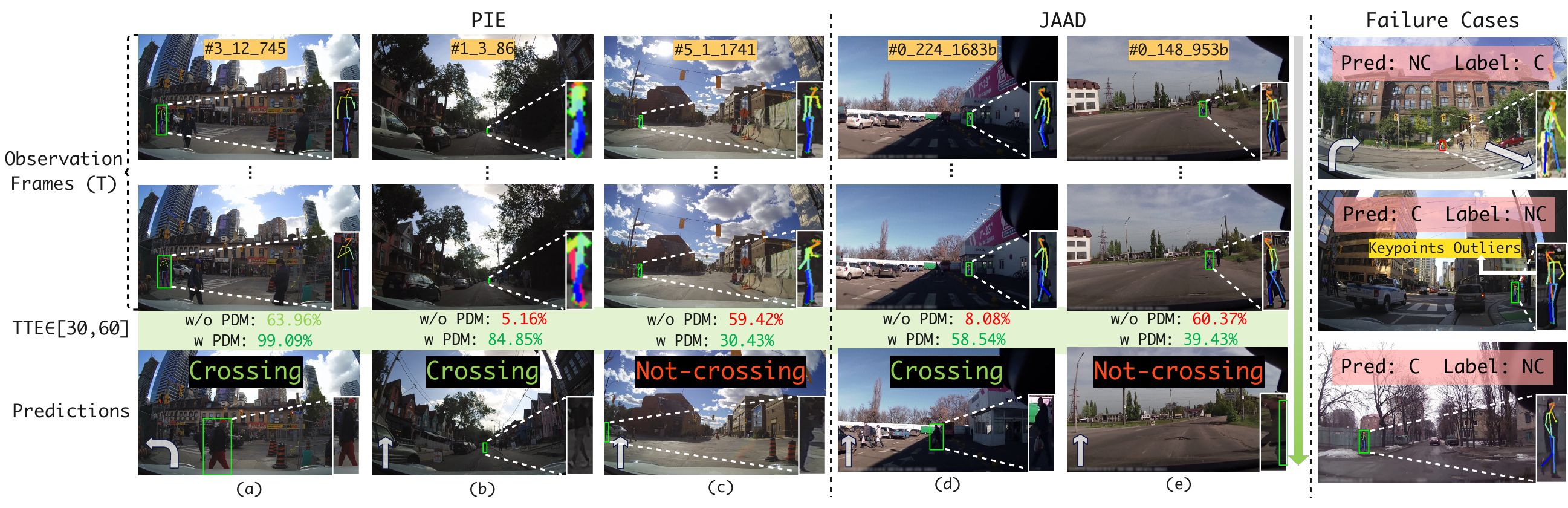}
  \vspace{-25pt}
  \caption{Visualizations of predicted pedestrian crossing intentions in challenging scenarios. Displayed are the crossing probabilities with and without our proposed PDM, where values above 50\% indicate crossing. Our method performs robustly under complex conditions, including irregular intersections (a) and curved roads (e), while also capturing long-range pedestrian behavior (b), and ambiguous roadside behavior (c) and (d). Failure cases typically falling into the parallel crossing behavior during vehicle turning, keypoints outliers, and the pedestrians yielding behavior to the ego-vehicle, as shown on the right.}
  \label{fig3}
\vspace{-20pt}
\end{figure*}

\begin{table}[t]
\caption{Performance comparisons of GTransPDM with state-of-the-art methods on the PIE and JAAD datasets. Best results in \textbf{bold}, while the second-best \underline{underlined}.}
\vspace{-8pt}
\label{tab1}
\resizebox{1.0\linewidth}{!}{
\begin{tabular}{l|cccc|cccc}
\hline
\multicolumn{1}{c|}{\multirow{2}{*}{Models}} & \multicolumn{4}{c|}{PIE}                                       & \multicolumn{4}{c}{JAAD}                                      \\
\multicolumn{1}{c|}{}                        & Acc$\uparrow$           & AUC$\uparrow$           & F1$\uparrow$            & P$\uparrow$             & Ac$\uparrow$           & AUC$\uparrow$           & F1$\uparrow$            & P$\uparrow$             \\
\hline
BiPed\cite{9711064}                                     & \underline{0.91} & \underline{0.90}          & \underline{0.85} & 0.82          & 0.83          & 0.79          & 0.60          & 0.52          \\
Pedestrian Graph+ \cite{9774877}                          & 0.89          & \underline{0.90} & 0.81          & 0.83          & \underline{0.86} & \underline{0.88} & 0.65          & \underline{0.58}          \\
\hline
MultiRNN \cite{8578539}                                   & 0.83          & 0.80          & 0.71          & 0.69          & 0.79          & 0.79          & 0.58          & 0.45          \\
SFRNN \cite{rasouli2020pedestrian}                                      & 0.82          & 0.79          & 0.69          & 0.67          & 0.84          & 0.84          & 0.65          & 0.54          \\
SingleRNN \cite{9304591}    & 0.81          & 0.75          & 0.64          & 0.67          & 0.78          & 0.75          & 0.54          & 0.44          \\
PCPA\cite{kotseruba2021benchmark}                                        & 0.87          & 0.86          & 0.77          & /             & 0.85          & 0.86          & \textbf{0.68} & /             \\
TrouSPI-Net\cite{gesnouin2021trouspi}                                 & 0.88          & 0.88          & 0.80          & 0.73          & 0.85          & 0.73          & 0.56          & 0.57          \\
IntFormer\cite{lorenzo2021intformer}                                   & 0.89          & 0.92          & 0.81          & /             & \underline{0.86} & 0.78          & 0.62          & /             \\
FF-STP\cite{9743954}                                   & 0.89          & 0.86          & 0.80          & 0.79             & 0.83 & 0.82          & 0.63          & 0.51             \\
PIT \cite{10247098}                                        & \underline{0.91} & \textbf{0.92} & 0.82          & 0.84          & 0.87          & \textbf{0.89} & \underline{0.67} & \underline{0.58}          \\
\hline
GTransPDM (w/o $X_{ke}$)                     & \textbf{0.92} & \underline{0.90} & \textbf{0.86} & \underline{0.85} & 0.84          & 0.72          & 0.54          & 0.56 \\
GTransPDM                              & 0.90          & 0.87          & 0.82          & \textbf{0.86} & \textbf{0.87} & 0.78          & 0.64          & \textbf{0.64} \\
\hline
\end{tabular}
}
\vspace{-10pt}
\end{table}

\begin{table}[t]
\centering
\caption{Model efficiency comparisons with recent models.}
\vspace{-8pt}
\label{tab2}
\resizebox{0.9\linewidth}{!}{
\begin{tabular}{c|cccc}
\hline
Model & 
\makecell{Acc \\ (PIE)} & 
\makecell{Acc \\ (JAAD)} & 
\makecell{Params \\ (Millions)} & 
\makecell{Inference \\ Time (ms)} \\ 
\hline
Pedestrian Graph+ \cite{9774877} & 0.89 & \underline{0.86} & \textbf{0.07} & 5.47 \\ 
PedGNN \cite{riaz2023synthetic} & 0.68 & 0.80 & -- & 0.58 \\ 
PCPA \cite{kotseruba2021benchmark} & 0.87 & 0.85 & 31.165 & 38.6 \\ 
PIT \cite{10247098} & \underline{0.91} & \textbf{0.87} & 21.2 & 4.8 \\ 
\hline
GransPDM (w/o $X_{ke}$) & \textbf{0.92} & 0.84 & \underline{0.13} & \textbf{0.03} \\ 
GransPDM & 0.90 & \textbf{0.87} & 0.23 & \underline{0.05} \\ 
\hline
\end{tabular}}
\vspace{-15pt}
\end{table}

\begin{table}[t]
\centering
\caption{The impact of reference line coordinates.}
\vspace{-8pt}
\label{tab3}
\resizebox{0.8\linewidth}{!}{
\begin{tabular}{c|ccc|ccc}
\hline
\multirow{2}{*}{$Y_A, Y_C$} & \multicolumn{3}{c|}{PIE} & \multicolumn{3}{c}{JAAD} \\
                         & Acc$\uparrow$   &AUC$\uparrow$     & F1$\uparrow$         & Acc$\uparrow$  &AUC$\uparrow$      & F1$\uparrow$        \\
\hline
$H$                        & 91.21   &88.13   & 81.61      & 87.43  & 78.16       & 64.00      \\
$H-100$                      & 91.24  & 87.70     & 81.42      & 87.25  & 77.81    & 63.46      \\
$H-200$                      & 91.16  &87.65    & 81.29      & 87.36  & 77.74       & 63.52      \\
\hline
\end{tabular}}
\vspace{-15pt}
\end{table}

\begin{table}[t]
\caption{Ablation study on multi-modalities. Ours in \textbf{bold}. Left: PIE dataset, Right: JAAD dataset.}
\vspace{-8pt}
\label{tab4}
\resizebox{1.0\linewidth}{!}{
\begin{tabular}{l|ccccc}
\hline
Encoders                    & \multicolumn{5}{c}{Choice}                                                                             \\
\hline
$X_{pe}$                          & \checkmark  &             &             & \checkmark  & \checkmark           \\
$X_{ev}$                          &             & \checkmark  &             & \checkmark  & \checkmark           \\
$X_{ke}$                          &             &             & \checkmark  &             & \checkmark           \\
\hline
\textit{\textbf{Acc}}$\uparrow$       & 88.17/84.76 & 84.73/82.52 & 82.19/84.45 & 90.43/84.43 & \textbf{91.21/87.43} \\
\textit{\textbf{AUC}}$\uparrow$       & 83.63/70.28 & 77.84/50.00 & 71.72/70.29 & 86.36/71.89 & \textbf{88.13/78.16} \\
\textit{\textbf{F1}}$\uparrow$        & 75.04/52.41 & 66.78/0.00  & 57.98/52.17 & 79.57/54.16 & \textbf{81.61/64.00} \\
\textit{\textbf{P}}$\uparrow$ & 75.08/57.71 & 68.94/0.00  & 65.81/56.42 & 80.53/55.82 & \textbf{80.97/64.11} \\
\textit{\textbf{R}}$\uparrow$ & 75.00/48.00 & 64.74/0.00 & 51.82/48.51 & 78.63/52.59 & 
\textbf{82.26/63.89} \\
\hline
\end{tabular}}
\vspace{-20pt}
\end{table}

\begin{table}[t]
\centering
\caption{Impact on features used in $X_{pe}$. $P_i(C)$: coordinates variation in PDM, $P_i(R)$: area ratio in PDM, $P_i(\Delta d)$: real depth variation. Best in \textbf{bold} and second-best \underline{underlined} in each dataset.}
\vspace{-8pt}
\label{tab5}
\resizebox{0.95\linewidth}{!}{
\begin{tabular}{l|ccccc}
\hline
$X_{pe}$                 & \multicolumn{5}{c}{Choice}                                          \\
\hline
$D_i$                & \checkmark       & \checkmark  & \checkmark  & \checkmark   & \checkmark        \\
$V_i$                      &              & \checkmark  & \checkmark  & \checkmark   & \checkmark        \\
$P_i(C)$                          &                          &             & \checkmark  & \checkmark  & \checkmark         \\
$P_i(R)$                 &             &                          &             & \checkmark      &     \\
$P_i(\Delta d)$                 &                          &             &             &       & \checkmark    \\
\hline
\textit{\textbf{Acc}}$\uparrow$       & 89.94/84.12 &  90.22/82.09 & 90.43/82.75 & \underline{91.21}/\textbf{87.43} & \textbf{91.44}/\underline{84.12}  \\
\textit{\textbf{AUC}}$\uparrow$       & 86.12/72.16 &  83.39/73.54 & 87.69/\underline{74.08} & \textbf{88.13/78.16} & \underline{88.12}/72.53  \\
\textit{\textbf{F1}}$\uparrow$        & 78.80/54.22 &  77.35/54.11 & 80.33/\underline{55.19} & \underline{81.61}/\textbf{64.00} & \textbf{81.93}/54.65  \\
\textit{\textbf{P}}$\uparrow$ & 78.76/\underline{54.66} &  \textbf{85.81}/49.00 & 78.30/50.57 & 80.97/\textbf{64.11} & \underline{82.01}/54.58  \\
\textit{\textbf{R}}$\uparrow$ & 78.85/53.78 &  70.41/60.41 & \textbf{82.48}/\underline{60.75} & \underline{82.26}/\textbf{63.89} & 81.85/54.72 \\
\hline
\end{tabular}}
\vspace{-20pt}
\end{table}

\vspace{-20pt}
\subsection{Evaluations}
The effectiveness of our method was evaluated using Accuracy, Area Under Curve (AUC), F1 Score, Precision (P), and Recall (R), which are commonly used in existing studies. Table \ref{tab1} and Table \ref{tab2} compare our method and efficiency with recent approaches from the past five years, including Transformer-based IntFormer~\cite{lorenzo2021intformer} and PIT~\cite{10247098}, the graph-based Pedestrian Graph+\cite{9774877}, and several RNN-based methods. Except for Pedestrian Graph+ and BiPed, others show the same configurations as ours, following \cite{kotseruba2021benchmark}. While PIT~\cite{10247098} captures traffic agent interactions using a ViT-based ~\cite{dosovitskiy2020image} network with 16 stacked Transformer layers, incurring high computational costs, our lightweight network achieves superior performance, reaching 0.92 accuracy and 0.86 F1 in PIE using only position and ego-vehicle motion encoders, with 0.86 precision when incorporating the pose encoder. However, since \cite{kotseruba2021benchmark} select samples following the sequence of long-period continuous videos, showcasing less randomness and performance degradation of $X_{ke}$ in our model. Thus, we conducted a 5-fold validation to further elucidate the contribution of $X_{ke}$ in Table \ref{tab6} of the Supplementary material and Table \ref{tab4}. Regarding JAAD, our method achieves an accuracy of 0.87 and a precision of 0.64, surpassing most existing approaches.

As an important component in the PDM, we evaluated the model's performance with different reference line coordinates (see Table~\ref{tab3}). Since the values $(\Delta x_t-\Delta x_{t-1}, \Delta y_t-\Delta y_{t-1})$ depend solely on the line's slope, we varied the vertical coordinates of points $A$ and $C$ during testing. Our results indicate that slight variations in the line slope have minimal impact on performance, thereby underscoring the robustness in enhancing lateral pedestrian movement recognition. To evaluate the effectiveness of each network component, ablation studies were conducted on the modalities and features used in $X_{pe}$, as shown in Table \ref{tab4} and \ref{tab5}. Left and right values represent metrics in PIE and JAAD respectively. In Table \ref{tab4}, the position encoder proves most effective in PIE, whereas skeletal information plays a critical role in JAAD. In contrast to the results in Table \ref{tab1}, incorporating skeletal information in PIE leads to a 1-2\% performance improvement. Overall, the model performs best when all encoders are combined. Table \ref{tab5} further illustrates the different features used in $X_{pe}$. Our proposed PDM brings 4.26\% and 9.89\% F1 improvements in PIE and JAAD respectively. To compare it with real depth-based methods, we computed the metric depth variation $\Delta d=d_t-d_{t-1}$ using Depth Anything~\cite{yang2024depth}, a SOTA monocular depth estimation model, as a substitute for $R$ in PDM. While the model achieves comparable performance on the PIE dataset, it shows a significant drop on JAAD. This is mainly due to the limitations of monocular depth estimation, such as sensitivity to occlusion and failure in scenes with poor visibility. Fig. \ref{fig3} presents challenging scenarios in PIE and JAAD datasets. Notably, in Fig.~\ref{fig3} (b), the ego-vehicle is traveling at high speed while the pedestrian remains at a distance, making it particularly challenging to detect the pedestrian's approach. Owing to our PDM, the model reliably identifies crossing behavior, thereby ensuring a safe response by AVs. Beyond behavior prediction, the core idea of our model also facilitates human behavior understanding in mobile sensing scenarios, such as capturing fine-grained pose changes and providing a lightweight alternative to depth sensing.

\vspace{-15pt}
\section{Conclusion}
\vspace{-5pt}

In this study, we propose GTransPDM, a \textbf{G}raph-embedded \textbf{Trans}former with a \textbf{P}osition \textbf{D}ecoupling \textbf{M}odule (PDM) for pedestrian crossing intention prediction. It effectively fuses multi-modal features with position, ego-vehicle motion, and pedestrian pose, ensuring a lightweight, real-time application. The PDM enhances pedestrian lateral movement recognition and captures depth cues, while a GCN-enhanced Transformer models spatio-temporal skeleton dynamics. Our approach predicts crossing intent 1–2 seconds in advance, achieving 92\% accuracy on PIE and 87\% on JAAD, with a runtime of 0.05ms, outperforming existing state-of-the-art methods. 

\bibliographystyle{IEEEtran}
\bibliography{ref}

\begin{thebibliography}{10}
\providecommand{\url}[1]{#1}
\csname url@samestyle\endcsname
\providecommand{\newblock}{\relax}
\providecommand{\bibinfo}[2]{#2}
\providecommand{\BIBentrySTDinterwordspacing}{\spaceskip=0pt\relax}
\providecommand{\BIBentryALTinterwordstretchfactor}{4}
\providecommand{\BIBentryALTinterwordspacing}{\spaceskip=\fontdimen2\font plus
\BIBentryALTinterwordstretchfactor\fontdimen3\font minus \fontdimen4\font\relax}
\providecommand{\BIBforeignlanguage}[2]{{%
\expandafter\ifx\csname l@#1\endcsname\relax
\typeout{** WARNING: IEEEtran.bst: No hyphenation pattern has been}%
\typeout{** loaded for the language `#1'. Using the pattern for}%
\typeout{** the default language instead.}%
\else
\language=\csname l@#1\endcsname
\fi
#2}}
\providecommand{\BIBdecl}{\relax}
\BIBdecl

\bibitem{8793991}
K.~Saleh, M.~Hossny, and S.~Nahavandi, ``Real-time intent prediction of pedestrians for autonomous ground vehicles via spatio-temporal densenet,'' in \emph{Proc. Int. Conf. Robot. Autom. (ICRA)}, 2019, pp. 9704--9710.

\bibitem{lin2024near}
C.~Lin, S.~Zhang, B.~Gong, and H.~Liu, ``Near-crash risk identification and evaluation for takeout delivery motorcycles using roadside lidar,'' \emph{Accid. Anal. Prev.}, vol. 199, p. 107520, 2024.

\bibitem{kotseruba2021benchmark}
I.~Kotseruba, A.~Rasouli, and J.~K. Tsotsos, ``Benchmark for evaluating pedestrian action prediction,'' in \emph{Proc. IEEE/CVF Winter Conf. Appl. Comput. Vis. (WACV)}, 2021, pp. 1258--1268.

\bibitem{9743954}
D.~Yang, H.~Zhang, E.~Yurtsever, K.~A. Redmill, and {\"U}.~{\"O}zg{\"u}ner, ``Predicting pedestrian crossing intention with feature fusion and spatio-temporal attention,'' \emph{IEEE Trans. Intell. Veh.}, vol.~7, no.~2, pp. 221--230, Mar. 2022.

\bibitem{10418196}
B.~Yang, J.~Zhu, C.~Hu, Z.~Yu, H.~Hu, and R.~Ni, ``Faster pedestrian crossing intention prediction based on efficient fusion of diverse intention influencing factors,'' \emph{IEEE Trans. Transport. Electr.}, pp. 1--1, Feb. 2024.

\bibitem{zhang2023trep}
Z.~Zhang, R.~Tian, and Z.~Ding, ``Trep: Transformer-based evidential prediction for pedestrian intention with uncertainty,'' in \emph{Proc. AAAI Conf. Artif. Intell. (AAAI)}, vol.~37, no.~3, 2023, pp. 3534--3542.

\bibitem{ahmed2023multi}
S.~Ahmed, A.~Al~Bazi, C.~Saha, S.~Rajbhandari, and M.~N. Huda, ``Multi-scale pedestrian intent prediction using 3d joint information as spatio-temporal representation,'' \emph{Expert Syst. Appl.}, vol. 225, p. 120077, 2023.

\bibitem{8876650}
Z.~Fang and A.~M. López, ``Intention recognition of pedestrians and cyclists by 2d pose estimation,'' \emph{IEEE Trans. Intell. Transp. Syst.}, vol.~21, no.~11, pp. 4773--4783, 2020.

\bibitem{10535049}
Y.~Ling, Z.~Ma, Q.~Zhang, B.~Xie, and X.~Weng, ``Pedast-gcn: Fast pedestrian crossing intention prediction using spatial–temporal attention graph convolution networks,'' \emph{IEEE Trans. Intell. Transp. Syst.}, pp. 1--14, 2024.

\bibitem{lorenzo2021intformer}
J.~Lorenzo, I.~Parra, and M.~Sotelo, ``Intformer: Predicting pedestrian intention with the aid of the transformer architecture,'' \emph{arXiv preprint arXiv:2105.08647}, 2021.

\bibitem{10161318}
A.~Rasouli and I.~Kotseruba, ``Pedformer: Pedestrian behavior prediction via cross-modal attention modulation and gated multitask learning,'' in \emph{Proc. IEEE Int. Conf. Robot. Autom. (ICRA)}, 2023, pp. 9844--9851.

\bibitem{9345505}
B.~Yang, W.~Zhan, P.~Wang, C.~Chan, Y.~Cai, and N.~Wang, ``Crossing or not? context-based recognition of pedestrian crossing intention in the urban environment,'' \emph{IEEE Trans. Intell. Transp. Syst.}, vol.~23, no.~6, pp. 5338--5349, 2022.

\bibitem{rasouli2020pedestrian}
A.~Rasouli, I.~Kotseruba, and J.~K. Tsotsos, ``Pedestrian action anticipation using contextual feature fusion in stacked rnns,'' in \emph{Proc. Brit. Mach. Vis. Conf.}, 2019, p. 171.

\bibitem{zhao2021action}
S.~Zhao, H.~Li, Q.~Ke, L.~Liu, and R.~Zhang, ``Action-vit: Pedestrian intent prediction in traffic scenes,'' \emph{IEEE Signal Process. Lett.}, vol.~29, pp. 324--328, 2021.

\bibitem{10493148}
K.~Kitchat, Y.-L. Chiu, Y.-C. Lin, M.-T. Sun, T.~Wada, K.~Sakai, W.-S. Ku, S.-C. Wu, A.~A.-K. Jeng, and C.-H. Liu, ``Pedcross: Pedestrian crossing prediction for auto-driving bus,'' \emph{IEEE Trans. Intell. Transp. Syst.}, vol.~25, no.~8, pp. 8730--8740, 2024.

\bibitem{9197434}
K.~D. Katyal, G.~D. Hager, and C.-M. Huang, ``Intent-aware pedestrian prediction for adaptive crowd navigation,'' in \emph{Proc. Int. Conf. Robot. Autom. (ICRA)}, 2020, pp. 3277--3283.

\bibitem{9827084}
L.~Achaji, J.~Moreau, T.~Fouqueray, F.~Aioun, and F.~Charpillet, ``Is attention to bounding boxes all you need for pedestrian action prediction?'' in \emph{Proc. IEEE Intell. Veh. Symp. (IV)}, 2022, pp. 895--902.

\bibitem{9304591}
I.~Kotseruba, A.~Rasouli, and J.~K. Tsotsos, ``Do they want to cross? understanding pedestrian intention for behavior prediction,'' in \emph{Proc. IEEE Intell. Veh. Symp. (IV)}, 2020, pp. 1688--1693.

\bibitem{dang2024coupling}
M.~Dang, Y.~Jin, P.~Hang, L.~Crosato, Y.~Sun, and C.~Wei, ``Coupling intention and actions of vehicle--pedestrian interaction: A virtual reality experiment study,'' \emph{Accid. Anal. Prev.}, vol. 203, p. 107639, 2024.

\bibitem{soares2021cross}
F.~Soares, E.~Silva, F.~Pereira, C.~Silva, E.~Sousa, and E.~Freitas, ``To cross or not to cross: Impact of visual and auditory cues on pedestrians’ crossing decision-making,'' \emph{Transp. Res. Part F Traffic Psychol. Behav.}, vol.~82, pp. 202--220, 2021.

\bibitem{kipf2016semi}
T.~N. Kipf and M.~Welling, ``Semi-supervised classification with graph convolutional networks,'' in \emph{Int. Conf. Learn. Represent. (ICLR)}, 2017.

\bibitem{vaswani2017attention}
A.~Vaswani, N.~Shazeer, N.~Parmar, J.~Uszkoreit, L.~Jones, A.~N. Gomez, L.~u. Kaiser, and I.~Polosukhin, ``Attention is all you need,'' in \emph{Adv. Neural Inf. Process. Syst. (NeurIPS)}, vol.~30, 2017.

\bibitem{8265243}
A.~Rasouli, I.~Kotseruba, and J.~K. Tsotsos, ``Are they going to cross? a benchmark dataset and baseline for pedestrian crosswalk behavior,'' in \emph{Proc. IEEE Int. Conf. Comput. Vis. Workshops (ICCVW)}, 2017, pp. 206--213.

\bibitem{rasouli2019pie}
A.~Rasouli, I.~Kotseruba, T.~Kunic, and J.~K. Tsotsos, ``Pie: A large-scale dataset and models for pedestrian intention estimation and trajectory prediction,'' in \emph{Proc. IEEE/CVF Int. Conf. Comput. Vis. (ICCV)}, 2019, pp. 6262--6271.

\bibitem{fang2022alphapose}
H.-S. Fang, J.~Li, H.~Tang, C.~Xu, H.~Zhu, Y.~Xiu, Y.-L. Li, and C.~Lu, ``Alphapose: Whole-body regional multi-person pose estimation and tracking in real-time,'' \emph{IEEE Trans. Pattern Anal. Mach. Intell.}, vol.~45, no.~6, pp. 7157--7173, 2022.

\bibitem{cox1958regression}
D.~R. Cox, ``The regression analysis of binary sequences,'' \emph{J. R. Stat. Soc. Ser. B Stat. Methodol.}, vol.~20, no.~2, pp. 215--232, 1958.

\bibitem{9711064}
A.~Rasouli, M.~Rohani, and J.~Luo, ``Bifold and semantic reasoning for pedestrian behavior prediction,'' in \emph{Proc. IEEE/CVF Int. Conf. Comput. Vis. (ICCV).}, 2021, pp. 15\,580--15\,590.

\bibitem{9774877}
P.~R.~G. Cadena, Y.~Qian, C.~Wang, and M.~Yang, ``Pedestrian graph +: A fast pedestrian crossing prediction model based on graph convolutional networks,'' \emph{IEEE Trans. Intell. Transp. Syst.}, vol.~23, no.~11, pp. 21\,050--21\,061, 2022.

\bibitem{8578539}
A.~Bhattacharyya, M.~Fritz, and B.~Schiele, ``Long-term on-board prediction of people in traffic scenes under uncertainty,'' in \emph{Proc. IEEE/CVF Conf. Comput. Vis. Pattern Recognit. (CVPR)}, 2018, pp. 4194--4202.

\bibitem{gesnouin2021trouspi}
J.~Gesnouin, S.~Pechberti, B.~Stanciulcscu, and F.~Moutarde, ``Trouspi-net: Spatio-temporal attention on parallel atrous convolutions and u-grus for skeletal pedestrian crossing prediction,'' in \emph{2021 IEEE Int. Conf. Autom. Face Gesture Recognit. (FG 2021)}.\hskip 1em plus 0.5em minus 0.4em\relax IEEE, 2021, pp. 01--07.

\bibitem{10247098}
Y.~Zhou, G.~Tan, R.~Zhong, Y.~Li, and C.~Gou, ``Pit: Progressive interaction transformer for pedestrian crossing intention prediction,'' \emph{IEEE Trans. Intell. Transp. Syst.}, vol.~24, no.~12, pp. 14\,213--14\,225, 2023.

\bibitem{riaz2023synthetic}
M.~N. Riaz, M.~Wielgosz, A.~G. Romera, and A.~M. L{\'o}pez, ``Synthetic data generation framework, dataset, and efficient deep model for pedestrian intention prediction,'' in \emph{2023 IEEE Int. Conf. Intell. Transp. Syst. (ITSC).}\hskip 1em plus 0.5em minus 0.4em\relax IEEE, 2023, pp. 2742--2749.

\bibitem{dosovitskiy2020image}
A.~Dosovitskiy, L.~Beyer, A.~Kolesnikov, D.~Weissenborn, X.~Zhai, T.~Unterthiner, M.~Dehghani, M.~Minderer, G.~Heigold, S.~Gelly, J.~Uszkoreit, and N.~Houlsby, ``An image is worth 16x16 words: Transformers for image recognition at scale,'' in \emph{Int. Conf. Learn. Represent. (ICLR)}, 2021.

\bibitem{yang2024depth}
L.~Yang, B.~Kang, Z.~Huang, X.~Xu, J.~Feng, and H.~Zhao, ``Depth anything: Unleashing the power of large-scale unlabeled data,'' in \emph{Proc. IEEE/CVF Conf. Comput. Vis. Pattern Recognit. (CVPR)}, 2024, pp. 10\,371--10\,381.

\end{thebibliography}

\newpage

\section{Supplementary material}
\subsection{The impact of $X_{ke}$ with K-fold validation.}

To further demonstrate the effectiveness of pose information in PIE, we conducted a 5-fold validation as detailed in Table~\ref{tab6}. Specifically, all pedestrian IDs in the PIE dataset were randomly shuffled and partitioned into five subsets. In each of the five iterations, four subsets (80\%) were used for training, while the remaining subset (20\%) was reserved for testing. Experiments were performed both with and without \(X_{ke}\), and the average values of the metrics were computed. Different from the benchmark select pedestrian samples following the sequence of long-period continuous videos, our approach provides a fairer comparison across each network component.

\vspace{-15pt}
\begin{table}[h]
\centering
\caption{The performance of $X_{ke}$ using 5-fold validation on PIE dataset.}
\vspace{-8pt}
\label{tab6}
\resizebox{0.9\linewidth}{!}{
\begin{tabular}{cc|ccccc}
\hline
Features & Fold & Acc$\uparrow$ & AUC$\uparrow$ & F1$\uparrow$ & P$\uparrow$ & R$\uparrow$ \\
\hline
\multirow{6}{*}{w/o \ $X_{ke}$} & 1 & 88.79\% & 86.78\% & 78.83\% & 75.27\% & 82.74\% \\ 
& 2 & 90.66\% & 86.86\% & 80.04\% & 80.39\% & 79.70\% \\ 
& 3 & 90.56\% & 86.67\% & 80.99\% & 83.40\% & 78.71\% \\ 
& 4 & 90.12\% & 87.09\% & 82.09\% & 84.18\% & 80.11\% \\ 
& 5 & 90.42\% & 87.88\% & 80.21\% & 77.49\% & 83.12\% \\ 
& \textbf{Avg.} & \textbf{90.11\%} & \textbf{87.06\%} & \textbf{80.43\%} & \textbf{80.15\%} & \textbf{80.88\%} \\ 
\hline
\multirow{6}{*}{w $X_{ke}$} & 1 & 90.14\% & 87.56\% & 80.82\% & 79.35\% & 82.34\% \\ 
& 2 & 91.97\% & 89.20\% & 83.09\% & 82.22\% & 83.97\% \\ 
& 3 & 91.23\% & 88.37\% & 82.78\% & 83.03\% & 82.53\% \\ 
& 4 & 90.98\% & 87.36\% & 83.21\% & 87.85\% & 79.03\% \\ 
& 5 & 91.47\% & 87.30\% & 81.31\% & 83.22\% & 79.49\% \\ 
& \textbf{Avg.} & \textbf{91.16\%} & \textbf{87.96\%} & \textbf{82.24\%} & \textbf{83.13\%} & \textbf{81.47\%} \\ 
\hline
\multicolumn{2}{c|}{\textit{\textbf{Improvement}}} & \textit{\textbf{1.05\%}} & \textit{\textbf{0.90\%}} & \textit{\textbf{1.81\%}} & \textit{\textbf{2.99\%}} & \textit{\textbf{0.60\%}} \\ 
\hline
\end{tabular}}
\end{table}

\vspace{-30pt}
\subsection{Impact on TTE variation and observation lengths.}
\begin{figure}[h]
  \centering
  \includegraphics[width=0.5\textwidth]{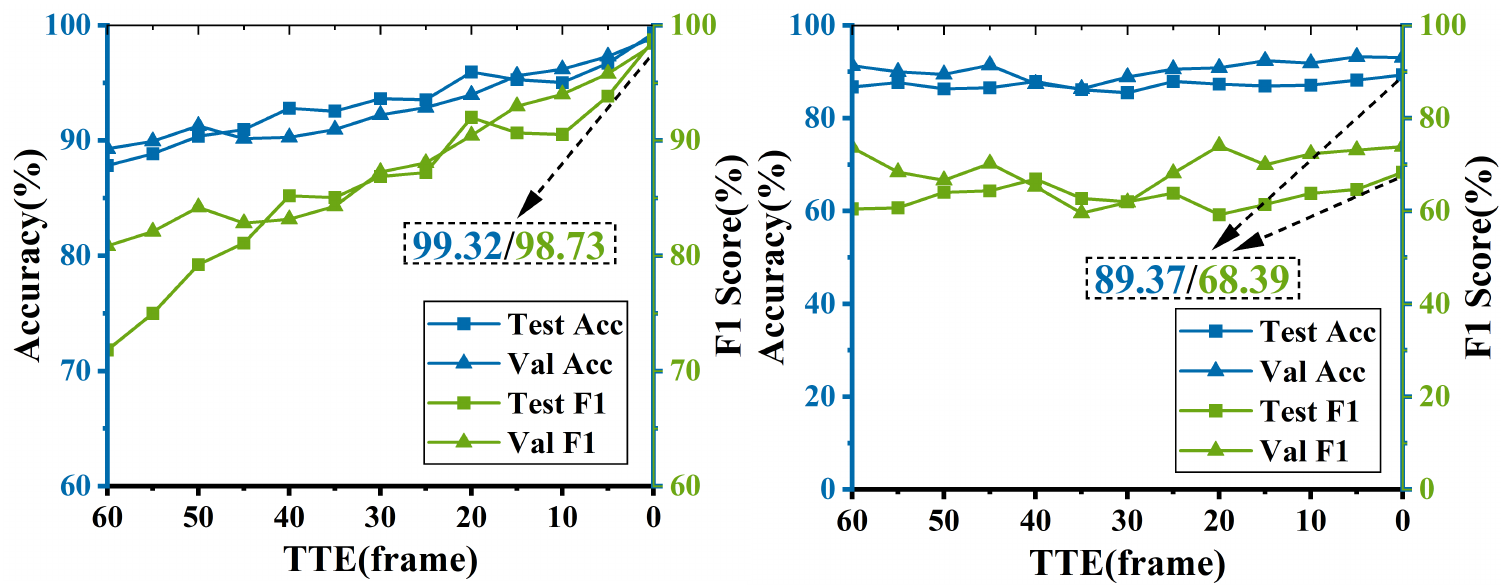}
  \vspace{-20pt}
  \caption{Performance evaluations on the PIE (left) and JAAD (right) datasets across various TTE settings in both validation and test sets.}
  \label{fig4}
\end{figure}

Fig.~\ref{fig4} shows the performance based on different TTE settings, ranging from 0 to 60 frames. A smaller TTE value indicates a shorter prediction interval. In the PIE experiment, we observed that as TTE decreased, the model's performance improved rapidly. Without a TTE interval (TTE = 0), the accuracy and F1 score reached 99.32\% and 98.73\%, respectively. However, in the JAAD experiments, the performance improved only when very close to the action start time, mostly because the JAAD\_all subset contains a large number of pedestrians who are far away and do not interact with the driver, which confused the model's ability to identify crossing behavior that was truly influenced by the ego-vehicle.

\vspace{-10pt}

\begin{figure}[h]
  \centering
  \includegraphics[width=0.48\textwidth]{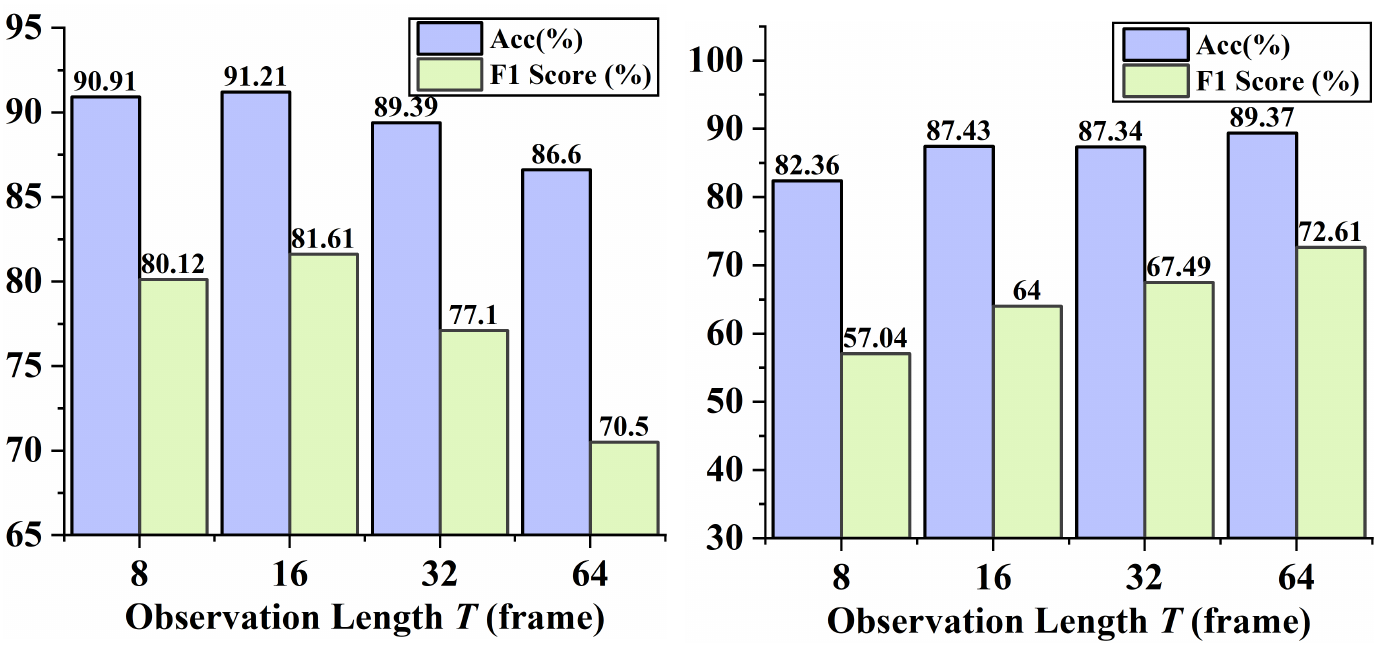}
  \vspace{-10pt}
  \caption{Impact of observation length on model performance across PIE (left) and JAAD (right) datasets.}
  \label{fig5}
\end{figure}

\vspace{-10pt}
Fig.~\ref{fig5} presents how the observation length influenced model performance. Remarkably, in the PIE dataset, prediction accuracy initially improved and then declined as the observation time extended, whereas in the JAAD dataset, longer observation times consistently resulted in higher accuracy. This is because the PIE contains more diverse pedestrian samples from urban downtown areas, where behavior before crossing can change unpredictably. As a result, extended observation times may lead to misjudgments by the model. In contrast, pedestrian crossing behavior in JAAD is relatively simpler, allowing longer observation periods to more effectively capture pedestrian movement patterns.

\vspace{-10pt}

\vspace{-25pt}
\subsection{Effect of motion features.}
In Table \ref{tab7}, we also evaluated the effectiveness of acceleration in PIE, resulting in a 2.43\% F1 improvement, with a particular increase in Recall. 

\vspace{-15pt}
\begin{table}[h]
\centering
\caption{Effect of motion features used in $X_{ev}$.}
\vspace{-8pt}
\label{tab7}
\resizebox{0.9\linewidth}{!}{
\begin{tabular}{cc|ccccc}
\hline
$S_i$         & $Acc_i$         & Acc$\uparrow$           & AUC$\uparrow$           & F1$\uparrow$            & P$\uparrow$     & R
$\uparrow$        \\
\hline
\checkmark    &               & 90.22\%       & 86.15\%       & 79.18\%       & 79.96\%       & 78.42\%       \\
              & \checkmark    & 90.60\%       & 85.52\%       & 79.29\%       & 83.04\%       & 75.85\%       \\
\checkmark    & \checkmark    & 91.21\%       & 88.13\%       & 81.61\%       & 80.97\%       & 82.26\%       \\
\hline
\multicolumn{2}{c|}{\textit{\textbf{Improvement}}} & \textit{\textbf{0.99\%}} & \textit{\textbf{1.98\%}} & \textit{\textbf{2.43\%}} & \textit{\textbf{1.01\%}} & \textit{\textbf{3.84\%}} \\
\hline
\end{tabular}}
\end{table}

\vspace{-25pt}
\subsection{Temporal encoder comparisons with RNNs.}

\begin{figure}[h]
  \centering
  \includegraphics[width=0.45\textwidth]{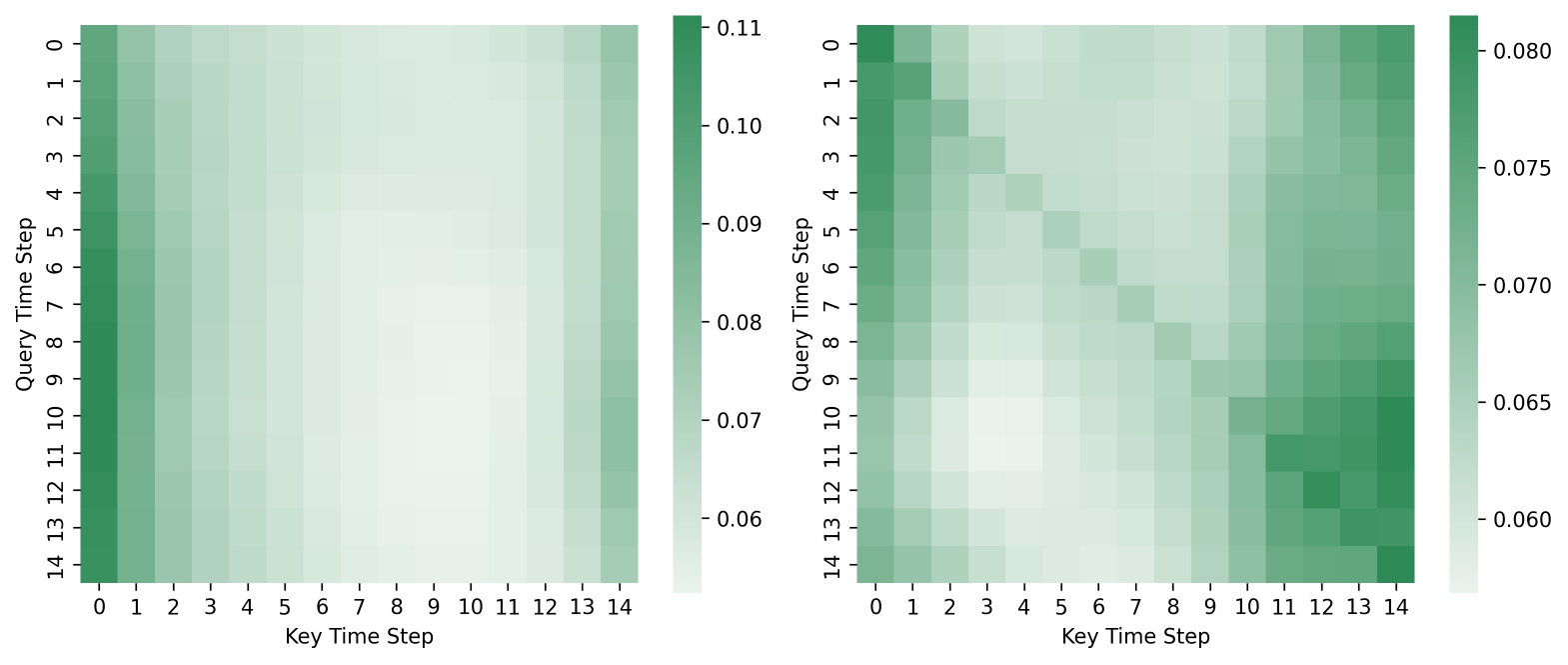}
  \vspace{-10pt}
  \caption{Temporal attention heatmap of PIE (left) and JAAD (right) datasets.}
  \label{fig6}
\end{figure}

Table~\ref{tab8} presents a comparison with RNN-based temporal encoders commonly used in PCIP. Benefiting from self-attention, our Transformer-based model achieves high accuracy and the best F1 scores on both datasets. As shown in Fig.~\ref{fig6}, the attention maps highlight distinct temporal patterns: in PIE, attention is concentrated on early time steps, indicating long-range dependency from the sequence start; in JAAD, each time step attends not only to itself, but also to both the beginning and end of the sequence.

\vspace{-15pt}
\begin{table}[h]
\centering
\caption{Temporal encoder comparisons with RNNs.}
\vspace{-8pt}
\label{tab8}
\resizebox{0.9\linewidth}{!}{
\begin{tabular}{l|ccc|ccc}
\hline
\multirow{2}{*}{Temporal Encoder} & \multicolumn{3}{c|}{PIE} & \multicolumn{3}{c}{JAAD} \\
 & Acc$\uparrow$ & AUC$\uparrow$ & F1$\uparrow$ & Acc$\uparrow$ & AUC$\uparrow$ & F1$\uparrow$ \\ 
\hline
LSTM & 90.08\% & \underline{86.83\%} & \underline{81.02\%} & \textbf{87.60\%} & 76.38\% & 62.51\% \\ 
GRU & \underline{90.78\%} & 86.78\% & 80.28\% & 86.47\% & \textbf{79.31\%} & 63.83\% \\ 
Transformer & \textbf{91.21\%} & \textbf{88.13\%} & \textbf{81.61\%} & \underline{87.43\%} & \underline{78.16\%} & \textbf{64.00\%} \\ 
\hline
\end{tabular}}
\vspace{-10pt}
\end{table}

\vspace{-10pt}

\vspace{-20pt}
\subsection{Impact of learnable edge importance.}

Table~\ref{tab9} illustrates the impact of incorporating learnable edge importance in GCN blocks. In this comparison, we evaluated the performance against the use of fixed edge weights. The results show that introducing learnable edge weights leads to marginal improvements.

\vspace{-15pt}
\begin{table}[H]
\centering
\caption{Performance impact of learnable edge importance.}
\vspace{-8pt}
\label{tab9}
\resizebox{0.9\linewidth}{!}{
\begin{tabular}{l|ccc|ccc}
\hline
\multirow{2}{*}{Edge weights} & \multicolumn{3}{c|}{PIE} & \multicolumn{3}{c}{JAAD} \\
 & Acc$\uparrow$ & AUC$\uparrow$ & F1$\uparrow$ & Acc$\uparrow$ & AUC$\uparrow$ & F1$\uparrow$ \\ 
\hline
Fixed edge & 91.06\% & 87.98\% & 81.60\% & 87.25\% & 78.06\% & 63.94\% \\ 
Learnable edge & \textbf{91.21\%} & \textbf{88.13\%} & \textbf{81.61\%} & \textbf{87.43\%} & \textbf{78.16\%} & \textbf{64.00\%} \\ 
\hline
\end{tabular}}
\vspace{-10pt}
\end{table}

\end{document}